\documentclass[runningheads]{llncs}
\usepackage[T1]{fontenc}
\usepackage{graphicx}

\usepackage{hyperref}
\usepackage{color}

\usepackage{amsmath}
\usepackage{listings}
\usepackage{microtype}
\usepackage{tabularx}
\usepackage{enumitem}
\usepackage{cite}

\begin{document}
\title{From Text to Space: Mapping Abstract Spatial Models in LLMs during a Grid-World Navigation Task}
\titlerunning{From Text to Space}
\author{Nicolas Martorell\inst{1,2}\orcidID{0000-0003-1778-7738}}
\authorrunning{N. Martorell}
\institute{Faculty of Exact and Natural Sciences, University of Buenos Aires, Argentina \and
National Scientific and Technical Research Council (CONICET) \\
\email{nmartorell@fbmc.fcen.uba.ar}\\
\url{https://exactas.uba.ar/}}
\maketitle              % typeset the header of the contribution
\begin{abstract}
Understanding how large language models (LLMs) represent and reason about spatial information is crucial for building robust agentic systems that can navigate real and simulated environments. In this work, we investigate the influence of different text-based spatial representations on LLM performance and internal activations in a grid-world navigation task. By evaluating models of various sizes on a task that requires navigating toward a goal, we examine how the format used to encode spatial information impacts decision-making. Our experiments reveal that cartesian representations of space consistently yield higher success rates and path efficiency, with performance scaling markedly with model size. Moreover, probing LLaMA-3.1-8B revealed subsets of internal units—primarily located in intermediate layers—that robustly correlate with spatial features, such as the position of the agent in the grid or action correctness, regardless of how that information is represented, and are also activated by unrelated spatial reasoning tasks. This work advances our understanding of how LLMs process spatial information and provides valuable insights for developing more interpretable and robust agentic AI systems.

\end{abstract}

\section{Introduction}

Large language models (LLMs) have demonstrated impressive capabilities in processing and generating text ~\cite{Brown2020,Bommasani2021}, yet a critical question remains: do these models develop abstract internal representations of the world, or do they simply memorize typical reasoning paths? This debate is particularly heated in non-textual domains—such as spatial or perceptual tasks—where the models have only been exposed to such concepts indirectly through text ~\cite{BenderKoller2020,Bisk2020,BenderEtAl2021}. While several studies have shown that LLMs can learn and solve tasks in these domains ~\cite{PatelPavlick2022,Wei2022,LiNyeAndreas2021}, and that their internal representations can be linearly mapped to those external structures ~\cite{GurneeTegmark2023}, a clear understanding of how these internal models are composed and how they influence behavior is still lacking.

This question is of increasing importance as LLMs are being deployed as agents that interact with environments by taking sequential actions based on textual inputs ~\cite{Yao2022,Wang2023}. In non-multimodal settings, the representation of spatial information as text is crucial, as language models are known to be highly sensitive to prompting ~\cite{Salinas2024,Leidinger2023,Anagnostidis2024}; the format in which this information is encoded may profoundly influence a model’s ability to extract relevant knowledge and make correct decisions. Furthermore, studying distinct representations of spatial information can help unveil if and how LLMs encode abstract models of space (i.e. internal activations that contain spatial information and are invariant to changes in prompting and context), and whether they leverage those internal world models to make decisions.

To address these issues, we investigated how LLM behavior and understanding of space change based on how spatial information is provided in the prompt, and whether there exists an internal model of space that is invariant to how that information is represented. We evaluated the LLaMA-3 family of models ~\cite{Dubey2024}, spanning model sizes from 1B to 90B parameters, on a Grid-World Spatial Orientation Task (GWSOT). In this task, models were required to navigate a 2D grid toward a goal by selecting one of four possible moves. We tested several different Spatial Information Representations (SIRs) in text form. We categorized these representations into three types: Cartesian representations, in which the (x, y) coordinates of both the agent and the goal are explicitly encoded; Topographic representations, which preserve the grid-like spatial structure in the text; and Textual representations, which describe the world state in prose. Across model sizes, we found that Cartesian representations consistently yielded better performance.

Furthermore, to gain insight into the internal processing of spatial information, we probed the activations of the LLaMA-3.1-8B model during the GWSOT. Our analysis revealed parameters that significantly predicted the position of the agent in the grid, as well as parameters that predicted action correctness in the subsequent step, regardless of how spatial information was represented in the prompt. Intriguingly, a subset of these units, primarily located in the model’s middle layers, were also more active when the model tackled spatial reasoning questions in unrelated contexts. These findings point to the existence of core units that form an internal spatial model invariant to prompt variations, context, and task specificity.

Our contributions can be summarized as follows:

\begin{samepage} % Prevents page breaks within the environment
\begin{itemize}[label={\scalebox{1}{$\bullet$}}, leftmargin=*] % Change the bullet to a dot
    \item We demonstrate that spatial orientation performance scales with model size.
    \item We show that specific ways of encoding spatial information have significant effects on model performance, even when the conveyed information is equivalent.
    \item We reveal that LLMs can represent the position of an agent in 2D space in ways that are partially invariant to prompting, with units that encode specific spatial features abstractly.
    \item We find specific units that predict action correctness during spatial navigation, which are also activated in unrelated spatial reasoning contexts.
\end{itemize}
\end{samepage}

\section{Related Work}

\subsubsection{World Models in LLMs.} World models have been defined as a compact, coherent, and interpretable representation of the generative process underlying the training data~\cite{GurneeTegmark2023,HaSchmidhuber2018,Vafa2024}. Some authors argue that LLMs lack internal world models capable of predicting world states and simulating action outcomes, which can impair their performance in agentic and planning tasks~\cite{Bisk2020,BenderEtAl2021}. On the other hand, some studies have identified specific neurons that encode space and time, demonstrating internal models that remain robust to variations in prompting~\cite{GurneeTegmark2023}. Other investigations have found evidence for internal models of the non-linguistic world—ranging from perceptual structures such as color to spatial orientation concepts, cardinal directions, and object properties~\cite{PatelPavlick2022,LiNyeAndreas2021,Abdou2021}. Implicit world models have also been described in goal-oriented contexts where the representation is influenced by an agent’s objectives~\cite{Li2021}.

\subsubsection{Internal Representations and Mechanistic Interpretability.} Mechanistic interpretability has emerged as a promising avenue for understanding the inner workings of LLMs~\cite{Olah2020,Elhage2021,Templeton2024}. This approach has been used to analyze emergent behaviors in larger models and to trace how internal representations influence output, providing essential insights for causality and AI safety~\cite{Wang2022,Nanda2023,Bereska2024}. For example, research has shown that neurons across different layers tend to specialize, with middle layers often containing neurons that represent higher-level contextual features~\cite{Geva2020,Durrani2022,Meng2022,Gurnee2023}. Furthermore, some neurons exhibit task-specific activations and can be predictive of model performance on these tasks~\cite{Meng2022,Wang2022b,LengXiong2024,Song2024}.

\subsubsection{Prompting Techniques and Prompt Influence on Outcome.} A growing body of work has established that LLM performance is highly sensitive to the specific prompting techniques employed, across a variety of benchmarks and tasks~\cite{Salinas2024,Leidinger2023,Anagnostidis2024}. Even variations in prompt formatting have been found to lead to notable differences in model behavior~\cite{Sclar2023}. This has driven the development of a wide range of prompting techniques which become highly relevant when attempting to improve LLM performance~\cite{Wei2022Chain,Sahoo2024}.

\subsubsection{LLMs for Spatial Navigation.} LLMs have also been applied to spatial navigation tasks despite being trained solely on text~\cite{Cote2019,Huang2022}. These tasks have commonly been represented as text-based sequences, where models are asked to provide information about the environment or actions that have effects in the world~\cite{Zhu2023,Yamada2023,Lin2023}. Previous studies have evaluated LLMs as agents navigating grid-world environments, where models must plan a full path in advance to locate goals and avoid obstacles~\cite{Aghzal2023,McDonald2023}. These approaches demonstrate that LLMs can solve spatial navigation tasks, although small models have shown limited generalization to variations such as differing grid sizes or obstacle configurations~\cite{Aghzal2023}.

\subsubsection{Spatial Maps in Neuroscience.} The neuroscience literature provides a foundational perspective on spatial representations through studies of spatial maps in the brain. Early work on place cells in the hippocampus of rats revealed that specific neurons encode an animal’s position in a two-dimensional space~\cite{OKeefeDostrovsky1971,OKeefe1976}. Subsequent research identified other spatially tuned cells—such as grid cells, which encode grid-like patterns; head-direction cells, which signal the direction of gaze; and boundary cells, which respond to environmental edges~\cite{Hafting2005,Lever2009,Taube1990}. The specialization observed in biological systems suggests that generalist artificial systems, including LLMs, might similarly develop internal representations for spatial orientation and navigation.

\section{Evaluating LLMs in a Spatial Orientation Task}

\begin{figure}[!t]
\centering
\includegraphics[width=\textwidth]{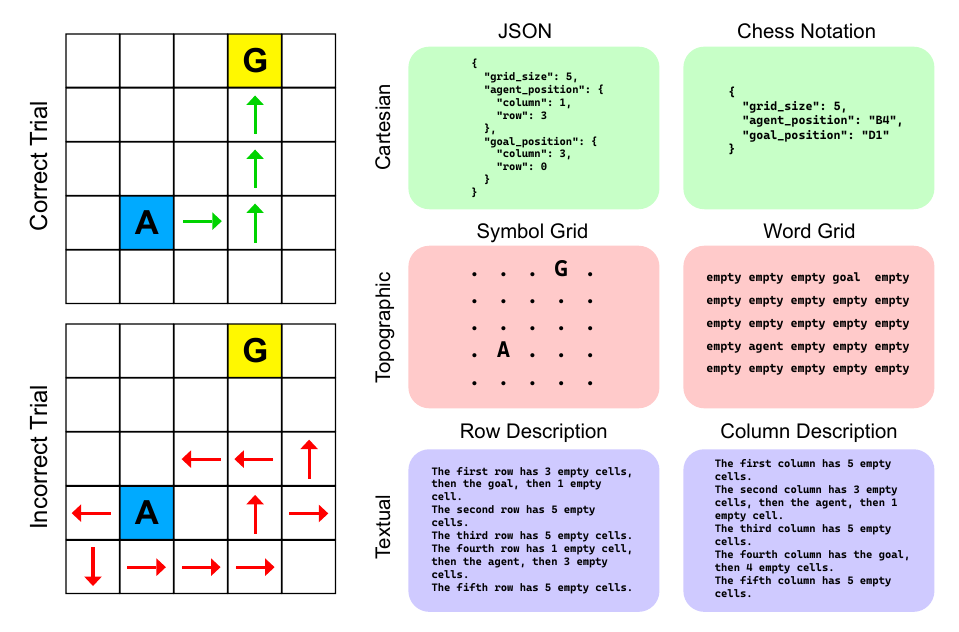}
\caption{The Grid-World Spatial Orientation Task (GWSOT). A Goal (G, yellow) is placed in a random position in a 5×5 grid (left-most panels). An Agent (A, blue) is placed in a semi-random location in the same grid, at least two steps away from the Goal. Green arrows (top panel) show a correct path which leads to the goal. Red arrows (bottom panel) shows an incorrect path which leads to task failure. The six panels on the right show examples of the three SIR classes and six SIR types used across this study.} \label{fig1}
\end{figure}

To evaluate spatial orientation in an agentic context, we designed the Grid-World Spatial Orientation Task (GWSOT), as illustrated in Fig.~\ref{fig1}. This is a simple variation of a classic reinforcement learning task~\cite{ChevalierBoisvert2018,ChevalierBoisvert2024}, where an agent and a goal are placed in an 5×5 grid. The two left-most panels of Fig.~\ref{fig1} depict example trials, where green arrows indicate a correct path leading to the goal, whereas red arrows show a path that result in failure by reaching a maximum number of steps.

At every step, the agent is provided with the current world state in a conversation-style prompt—where user messages describe the grid state and assistant messages record previous decisions. The LLM is required to respond with a JSON object containing a single key, "action", whose value is one of four commands: "UP", "DOWN", "LEFT", or "RIGHT". An action is defined as correct if it reduces the Manhattan distance between the agent and the goal. A trial ends when the agent reaches the goal or when it exceeds a maximum step limit.

We use different Spatial Information Representations (SIRs) to convey the current world state to the LLM. We categorized SIRs into three classes:

\begin{samepage} % Prevents page breaks within the environment
\begin{itemize}[label={\scalebox{1}{$\bullet$}}, leftmargin=*] % Change the bullet to a dot
    \item Cartesian representations: The LLM receives explicit (x, y) coordinates for the agent and the goal, along with the grid size.
    \item Topographic representations: The two-dimensional structure of the grid is preserved by representing cells and objects using characters or words. This requires LLMs to be able to interpret the text layout as structure, which previous research has shown these models are able to achieve~\cite{Li2024}.
    \item Textual representations: The world state is described in prose-like language.
\end{itemize}
\end{samepage}

To ensure that our findings are attributable to the SIR class rather than to a specific formatting style, we implemented two variants for each class: JSON and Chess Notation for Cartesian, Symbol and Word Grid for Topographic, and Row and Column Description for Textual (see Fig.~\ref{fig1}, right panels). Despite their differences, all six SIR types encode identical spatial information.

For additional details on the task configuration and system prompt see Appendix A.

\section{Prompting and Scale as Factors in Performance}

We evaluated the LLaMA 3 family of models—specifically the 3.2-1B, 3.2-3B, 3.1-8B, 3.2-11B, 3.1-70B, and 3.2-90B variants—in the 5×5 GWSOT. For each model and each of the six SIR types, we conducted 100 trials, resulting in a dataset of 3,600 trials. In addition, we ran 100 trials with a random policy agent that uniformly selected one of the four possible actions at each step, providing a baseline for comparison. A complete account of each experiment performed in this work, along with their features and number of trials per condition can be found in Appendix B.

Model performance was quantified using three metrics. First, we computed the success rate as the proportion of trials in which the agent reached the goal. Second, for successful trials, we measured path efficiency by dividing the minimum number of steps required to reach the goal (i.e., the initial Manhattan distance) by the actual number of steps taken by the agent; an efficiency of 1 indicates a perfect, direct path. Third, for unsuccessful trials, we calculated the final distance ratio by dividing the final Manhattan distance from the agent to the goal by their initial Manhattan distance. A ratio of 1 implies no improvement, while lower ratios indicate that the agent moved closer to the goal despite not reaching it.

\begin{figure}[!t]
\centering
\includegraphics[width=\textwidth]{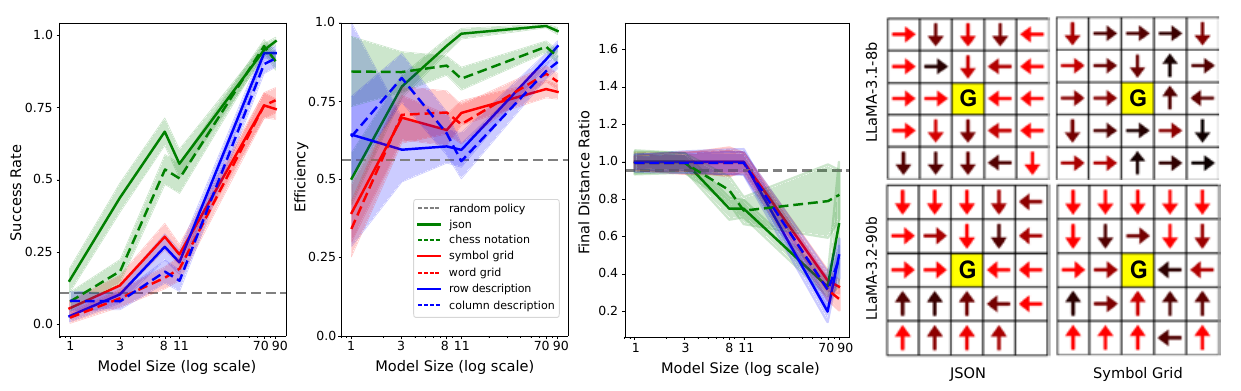}
\caption{Model performance improves with model size and is influenced by SIR type. The first three panels show performance metrics in the 5×5 GWSOT for Cartesian (green), Topographic (red) and Textual (blue) SIR classes. In all cases, the gray dashed line shows the performance of a random policy agent. Left-most panel shows the success rate of LLMs in this task as a function of model size. Second panel show the mean efficiency of LLMs in this task as a function of model size, only for trials where the goal was reached. Third panel shows the mean final distance ratio as a function of model size, only for trials where the goal was not reached. In all cases, the x axis (model size) is logarithmic and shadings represent standard error. The right-most four panels depict example policy maps for the LLaMA-3.1-8B and the LLaMA-3.2-90B models, for the JSON and Symbol Grid SIRs. Arrows represent the most common action chosen by the model in each relative position to the goal. Brighter red arrows represent more commonly taken actions, whereas darker arrows represent uncertain decisions.} \label{fig2}
\end{figure}

The first three panels of Figure~\ref{fig2} summarize how both model scale and SIR type influence these performance metrics. The left panel shows that success rates increase with model size for every SIR type (Linear Regression, $\beta = 0.008$, $p < .001$). The smallest models (1B and 3B) perform near chance level---around 10\%---for all SIR types except for JSON, while the largest models (70B and 90B) always exceed a 74\% success rate. Particularly, the 90B model receiving the JSON representation achieved an outstanding 98\% success rate. Notably, Cartesian SIRs consistently outperformed Topographic and Textual SIRs across model sizes (Binomial GLM: success $\sim$ representation + model\_size, $\beta_{\text{topographic}} = 1.385$, $p < .001$; $\beta_{\text{textual}} = 1.270$, $p < .001$), although the mid-sized models (8B and 11B) exhibited the most pronounced performance differences between SIRs. For example, the 8B model achieves a 66\% success rate with the JSON SIR but only 30\% with its best non-cartesian SIR (Symbol Grid). Importantly, this pattern holds regardless of formatting variations within each SIR class (compare same-color curves in the left panel of Fig.~\ref{fig2}---both JSON and Chess Notation outperform Topographic and Textual SIRs).

A similar trend was observed in the efficiency metric, shown in the second panel of Fig.~\ref{fig2}. This metric also improved with the scale of the model (Gaussian GLM: efficiency $\sim$ model\_size, $\beta_{\text{model\_size}} = 0.001$, $p < .001$), and Cartesian SIRs consistently resulted in models taking more efficient paths to the goal during successful trials (Gaussian GLM: efficiency $\sim$ representation + model\_size, $\beta_{\text{topographic}} = -0.159$, $p < .001$; $\beta_{\text{textual}} = -0.116$, $p < .001$), particularly in the mid-sized models where performance is above chance and not yet saturated. Notably, the largest models approached near-optimal efficiency in this task when receiving the JSON SIR (see green continuous curve).

Furthermore, the final distance ratio (third panel of Fig.~\ref{fig2}) also improves with scale (Gaussian GLM: final\_distance\_ratio $\sim$ model\_size, $\beta_{\text{model\_size}} = -0.006$, $p < .001$). Smaller models (1B and 3B) fail to improve their positions in unsuccessful trials regardless of SIR type, while in mid-sized models (8B and 11B) only Cartesian representations led to improvements (Welch’s t-test; Cartesian: $t = -1.96$, $p = 0.053$, marginally significant; Topographic: $t = 1.06$, $p = 0.291$; Textual: $t = 1.41$, $p = 0.162$). For the largest models (70B and 90B), all SIR types demonstrated improvements relative to the random policy (Welch’s t-test; Cartesian: $t = -3.055$, $p = 0.004$; Topographic: $t = -9.223$, $p < 0.001$; Textual: $t = -7.324$, $p < 0.001$). Overall, these metrics reveal a scaling law for spatial navigation performance, and show that Cartesian representations are better tuned to convey spatial information to the models.

To gain further insight into the models’ spatial reasoning, we constructed policy maps—the four right-most panels of Fig.~\ref{fig2}—where cells represent positions relative to the goal (recentered to the middle of the grid to allow visualization), and arrows indicate the most common action chosen in each cell, colored by the relative frequency with which that action was chosen at each position in the grid. For brevity, only maps for the LLaMA-3.1-8B and LLaMA-3.2-90B models using JSON and Symbol Grid SIRs are shown. Presenting the models with a Cartesian SIR resulted in a higher proportion of correct policies across positions and model sizes, compared to other SIR types. However, larger models found the correct policy in more positions and were more consistent in their behavior. This aligns with the quantitative performance metrics, reinforcing the conclusion that Cartesian representations (and especially the JSON SIR) lead to more reliable spatial decision-making.

\section{Activations Encode Spatial Features in a Mid-Sized LLM}

To investigate whether specific parameters within LLMs are involved in representing spatial information, we conducted a detailed analysis of the LLaMA-3.1-8B model—the smallest variant that demonstrated above-chance performance across all SIR types (see left-most panel of Fig.~\ref{fig2}). Our primary aim was to determine whether activations in individual layers encode the grid configuration up to a linear transformation and, if so, what are the components of that spatial model and how it depends on the SIR type.

\begin{figure}[!t]
\centering
\includegraphics[width=\textwidth]{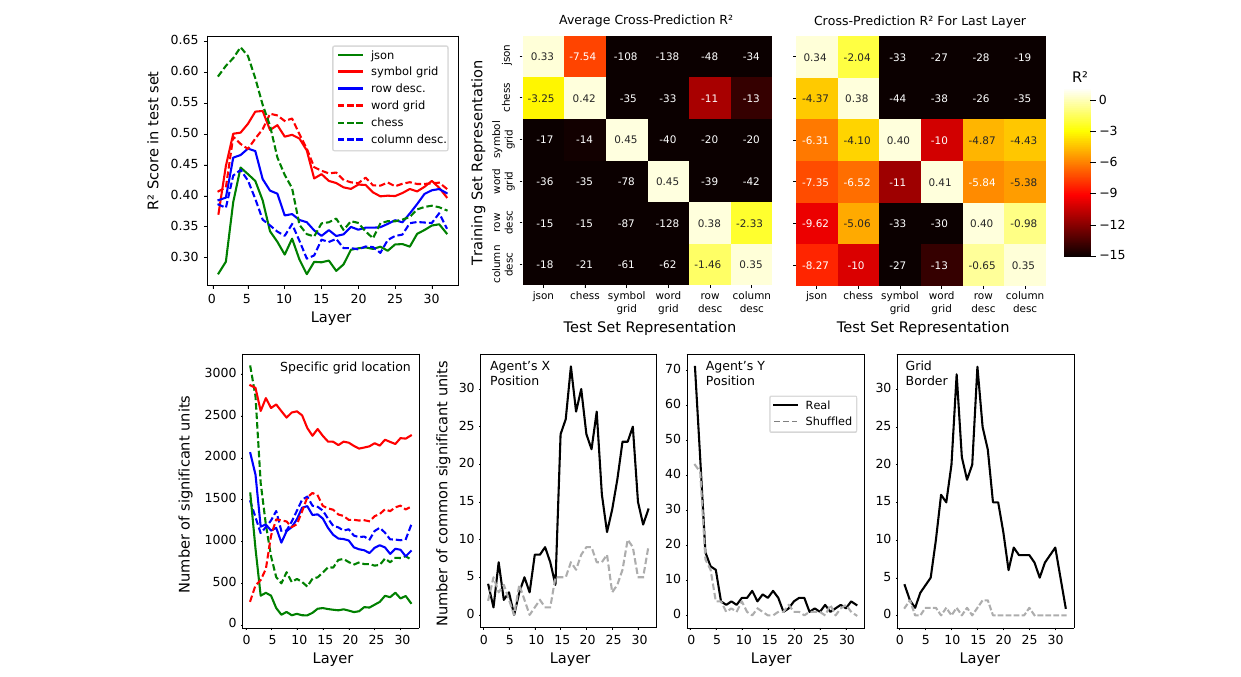}
\caption{Activations in LLaMA-3.1-8B predict grid configuration. Left-most top panel shows the R² of linear models trained on activations from individual layers to predict the full configuration of the 5×5 grid for each of the six SIR types. Second top panel shows the R² of linear models trained on each individual SIR type (rows) and evaluated on every individual SIR type (columns), averaged across models trained on each layer. The color scale is cut-off at -15 for visualization purposes. The third top panel shows the same cross-prediction R² measure but evaluated only for models trained on the last layer. The left-most bottom panel shows the number of parameters from each layer that were significantly correlated with the agent’s position being in one specific cell from the 5×5 grid, for each SIR type. The last three bottom panels show the number of units in each layer that were significantly correlated with the agent’s x position, y position or with border cells, for all six SIR types (black curves). Dashed gray lines shows the same calculation performed on shuffled parameter indices.} \label{fig3}
\end{figure}

We then trained linear regression models to predict the complete configuration of a 5×5 grid, represented as a 50‑dimensional binary vector (with 25 dimensions encoding the presence or absence of the agent and 25 for the goal—details on this analysis can be found on Appendix C). The R² values of these linear models generally peaked in early middle layers and then declined in later layers, as depicted in the left‑most top panel of Fig.~\ref{fig3}. This was the case for all SIR types. However, the layer at which the R² peaked depended on the SIR class. For Cartesian and Textual representations, R² peaked around layer 5, exhibited a dip, and then showed a modest recovery in the final layers. In contrast, Topographic representations reached their maximum performance deeper in the model (around layer 10) before declining monotonically. These observations suggest that the way the model represents spatial information, and how this representation changes across layers, depends on how that information is encoded in text. Despite these differences, the test‐set performance of every layer for each SIR type consistently exceeded that of a null model (Permutation Test: one per model against 10 shuffles, p < .001 for all cases), indicating that information about the spatial configuration of the grid is present throughout the network regardless of SIR type.

To assess the generalizability of these internal representations, we performed cross‑SIR predictions by training linear models on activations derived from one SIR type and evaluating its performance on activations from the others. Averaging R² scores across layers revealed that representations do not generalize fully between different SIR types—performance dropped markedly when the training and testing SIRs differed (top middle panel in Fig.~\ref{fig3}). However, this drop was considerably smaller when evaluating within the same SIR class (for example, when training on JSON and testing on Chess Notation, or between different Textual variants). As the variance of R² values between SIR types decreased from superficial to deep layers (left-most top panel of Fig.~\ref{fig3}), we wondered if grid representations might also become more consistent in deeper layers. Motivated by this hypothesis, we examined cross-SIR performance using only the final layer activations (right-most top panel in Fig.~\ref{fig3}). Indeed, we found that R² values were generally larger for models trained on this layer than in the average case. Furthermore, the best performance for every SIR was achieved when training and testing on the same SIR type, with the second‑best performance always occurring for the other member of the same SIR class. However, only R² values for same-SIR testing were found to be significantly larger than those obtained by evaluating a null model (Permutation Test: one per comparison against 10 shuffles, p <0.05 only for models tested on same-SIR activations). These findings suggest that the representation of grid space is mostly specific to each SIR, but there appears to be some component of that representation that is generalizable at least between SIRs of the same class. Based on this observation, we turned to look for abstract components of the model that could be representing spatial information irrespective of SIR type.

Inspired by the existence of place cells in biological systems, which activate only when an animal is in a specific position in space~\cite{OKeefeDostrovsky1971,OKeefe1976}, we next explored whether individual units in the model exhibit spatial selectivity with respect to the agent’s position in the grid. For each SIR type, we fitted a linear model per individual parameter, using activations as regressors for predicting whether the agent occupied a specific cell of the 5×5 grid. We performed this analysis for all model parameters and for all 25 grid cells, and adjusted p-values by the number of comparisons using the Bonferroni correction (see Appendix C for more details). We found hundreds to thousands of units per layer (depending on the SIR type—see left-most bottom panel in Fig.~\ref{fig3}) that were significantly correlated with the agent’s presence in specific grid cells. However, these correlations might reflect the encoding of particular textual motifs within a SIR rather than a true, abstract spatial model. To attempt to find units that encode spatial information in a general way, irrespective of how that information is represented in the input, we looked for parameters that showed a correlation with a specific grid cell across all SIR types. Interestingly, we found no such units, suggesting that the model does not represent the position of the agent by recognizing specific spatial locations, as is the case in biological systems.

We then wondered if the agent’s position in the grid might instead be encoded continuously in the x and y axes. To evaluate this, we looked for units whose activations were correlated with the agent’s x or y coordinates. We found 448 parameters that were significantly correlated with the agent’s x position across all SIR types, predominantly in intermediate and deeper layers. Shuffling the parameter indices between SIR types markedly reduced the number of common significant units (compare black and gray curves in the bottom second panel of Fig.~\ref{fig4}, Wilcoxon signed-rank test, W = 7, p < .001). In contrast, units encoding the agent’s y coordinate regardless of SIR type were less common but still existed (258 in total) and tended to be concentrated in the most superficial layers (bottom third panel, Wilcoxon signed-rank test, W = 27, p < .001). Additional information on these analyses can be found in Appendix D.

Finally, motivated by the concept of boundary cells in neuroscience, which are activated when an animal is close to an environment’s boundary~\cite{Lever2009}, we next looked for units which consistently encoded whether the agent was located in a border cell. We identified 373 units that were significantly correlated with this condition for all SIR types—far exceeding the number observed in a shuffled control (compare black and gray curves in the right‑most bottom panel of Fig. 4, Wilcoxon signed-rank test, W = 2, p < .001). These invariant units were primarily concentrated in intermediate layers (approximately layers 8–18).

Overall, our findings indicate that the LLaMA‑3.1‑8B model builds an internal representation of grid space that is partially dependent on how spatial information is represented, but also contains components that show a high level of abstraction in encoding spatial features such as the location of the agent in 2D space.

\section{Robust Units Encode Spatial Reasoning Across Unrelated Tasks}

Spatial reasoning does not only involve representing locations in space—it requires detecting relations between objects’ positions to make predictions and guide decision-making. For an LLM to effectively navigate its environment, it must be able to understand whether its actions will move it closer to or farther from its target. This capability—predicting action correctness—requires an internal representation of the relative positions of objects within the environment. With this in mind, we next investigated whether the LLaMA‑3.1‑8B model exhibits this form of spatial reasoning representations.

\begin{figure}[!t]
\centering
\includegraphics[width=\textwidth]{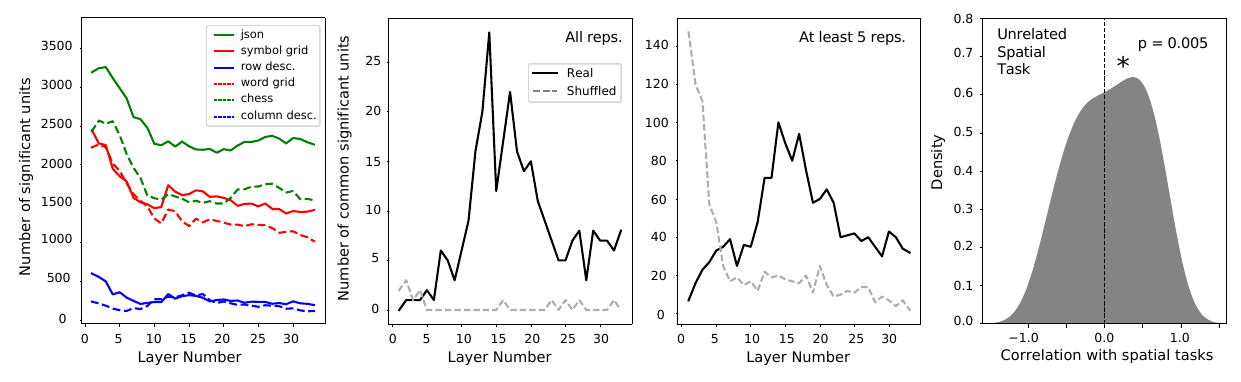}
\caption{Activations in LLaMA-3.1-8B predict action correctness across representation types. The left-most panel compares the total number of parameters per layer that are significantly correlated with action correctness for each SIR type. The second and third panels show the number of common significant parameters across all representation types and across at least 5 of the 6 representation types, respectively (black curves) compared to a shuffle between representations (gray dashed curves). The last panel shows the distribution of correlation coefficients between parameter activation and the spatial nature of a question in an unrelated task, for each parameter identified in the second panel as significantly correlated with action correctness for all SIR types in the GWSOT.} \label{fig4}
\end{figure}

We again fitted a linear model per parameter, this time using activations during the generation of the action keyword as predictors of action correctness (i.e., whether a selected action moves the agent closer to the goal) and applying a Bonferroni correction for the total number of parameters in the model. Our results revealed that all model layers contained hundreds to thousands of parameters that were significantly correlated with action correctness (Fig.~\ref{fig4}, first panel), paralleling results from the spatial features studied in the previous section. Interestingly, both Cartesian representations—and especially the JSON variant—yielded the highest number of significant parameters, followed by Topographic and then Textual representations. The number and distribution of these parameters was quite consistent between SIRs of the same class (compare curves of the same color in first panel), reinforcing the fact that differences between SIRs arise from the spatial encoding method rather than from specific prompt formatting. Interestingly, the parameter count for the JSON SIR was slightly higher than for the Chess Notation SIR, mirroring prior results from model’s behavior (Fig.~\ref{fig2}, compare green curves in the first three panels—success rate, efficiency and final distance ratio tend to be improved for JSON with respect to Chess Notation). This suggests that prompt formatting could be playing a smaller but significant role in the Cartesian case.

We then asked if the model contained units that showed this correlation with action correctness irrespective of how spatial information was represented. Indeed, we identified a subset of 286 units that were significantly correlated with action correctness for all six SIR types (Fig.~\ref{fig4}, second panel), far exceeding the 12 units found in a shuffled control (Wilcoxon signed-rank test, W = 12.5, p < .001). These 286 units were predominantly located in the middle layers of the model (approximately layers 12–22), consistent with the distribution of units encoding the agent’s position (Fig.~\ref{fig3}) and with previous studies reporting that higher-level conceptual processing in LLMs tends to emerge in intermediate layers~\cite{Geva2020,Durrani2022,Meng2022,Gurnee2023}.

Given that this model had low performance in the GWSOT for some SIR types (i.e. both the Column Description and Word Grid SIRs had success rates below 20\%), we considered that our inclusion criterion for this core group of parameters might be excluding units that nevertheless encode abstract spatial information. To explore this, we ran the same analysis using a more permissive threshold, identifying units that were significantly correlated with action correctness in at least five of the six SIR types. Under this relaxed criterion, we identified 1,560 units across all model layers (Fig.~\ref{fig4}, third panel; there are more common units than expected by chance: Wilcoxon signed-rank test, W = 81.5, p < .001). Notably, even with this substantial increase in the number of detected parameters, their distribution remained strikingly consistent: most of these units were concentrated in the middle layers, and almost none were located in earlier layers where most of the significant parameters for single SIRs are found.

These findings, together with our results from Figure~\ref{fig3}, support the presence of a genuine spatial information encoding system within the model, characterized by a concentration of abstract, prompt-independent units in its middle layers. However, if these units are really functioning as a spatial world model, we would expect them to be involved in spatial reasoning regardless of the specific task. To test this, we examined the activation of the 286 core action-correctness encoding units in a different context. We presented the LLaMA-3.1-8B model with a set of spatial-reasoning and non-spatial-reasoning questions (200 each, examples in Appendix E) and averaged the activations of these core units across all input tokens. For each unit, we computed the correlation between its average activation and a binary indicator of whether the question was spatial or not. The resulting distribution of correlation coefficients was significantly higher than zero (Fig.~\ref{fig4}, right-most panel, Wilcoxon signed-rank test, W = 16634, p = 0.005), indicating that these units are more active when receiving spatial-reasoning questions than when receiving questions that are not conveying information about space. In contrast, randomly chosen units—matching the number of core units per layer—did not exhibit this bias (Wilcoxon signed-rank test, W = 19108, p = 0.31). This suggests that these parameters, which predict action correctness in the GWSOT, are indeed processing spatial information in a task-independent way. Interestingly, when we performed the same analysis for the core units that encode the agent's position in the 2D space, correlations were not significantly higher than 0 (Wilcoxon signed-rank test, W = 129291, p = 0.78), supporting the hypothesis that abstract spatial reasoning is related with inner representations of outcome prediction rather than with the encoding of specific spatial features such as object locations.

Finally, we asked if silencing these core spatial units would decrease the performance of the model when navigating our spatial orientation task. To evaluate this, we conducted an ablation experiment in which we silenced these 286 neurons during the 5×5 GWSOT using the JSON SIR (more information on ablation experiments can be found in Appendices B and C). Interestingly, there was almost no impact in the performance of the model: it achieved 55\% success rate during ablation trials compared to 59\% for the intact model. The same was true when silencing units correlated with the agent’s position for all SIR types (those found in Fig.~\ref{fig3}): performance in this task was not reduced (62\%). This suggests that abstract spatial units constitute only a minor component of the overall spatial encoding system in LLMs, and that SIR-specific units are enough to solve this task. To evaluate if SIR-specific units were indeed necessary, we attempted to test the model while ablating all units significantly correlated with action correctness for the JSON SIR. However, these comprised 59.7\% of the total parameter count across all layers (see green curve in left-most panel of Fig.~\ref{fig4}; also a discussion of this point in Appendix D). Consequently, the model was unable to output coherent actions in these trials. This was the case even when we restricted parameter ablation to the middle layers (12-22). Our results suggest that spatial information is widely distributed throughout the model, likely comprised mostly of polysemantic neurons that cannot easily be perturbed without disrupting general model behavior and output coherence.

\section{Discussion}

Our results demonstrate that representing spatial information in Cartesian coordinates yields the best performance for LLMs in the GWSOT. Cartesian representations consistently outperform Topographic and Textual SIRs across a range of model sizes. It’s interesting, but perhaps expected, that LLMs prefer to process spatial information when it’s presented mathematically, where the x and y dimensions can be operated separately to find the correct orientation, instead of relying on an intuitive understanding of topographical space like humans would. It is also interesting that the JSON variant usually outperforms the Chess Notation SIR, which could be due to the prevalence of such representations in the training data.

We designed a very simple spatial orientation task to be able to test the spatial reasoning of smaller models. While this setting allowed us to isolate the effects of different spatial representations, it also limits the generalizability of our findings to more complex spatial reasoning scenarios. Future work should extend these experiments to larger models and more challenging tasks, such as navigating environments with obstacles, multiple goals, or even three-dimensional spaces. Such studies would help determine whether the advantages of representing space in certain ways persist under more demanding conditions usually found in real-world agentic tasks.

Another promising direction is the exploration of multimodality in spatial reasoning. Currently, our study compares distinct text-based representations of spatial information. Recent works have investigated how combining image and text prompts can impact performance in spatial reasoning tasks~\cite{Wang2024,Ranasinghe2024}. Future work should further explore how integrating visual cues with specific textual descriptions of space might provide complementary advantages in agentic frameworks.

Our analysis of the LLaMA-3.1-8B model provides compelling evidence that even relatively small LLMs develop abstract models of spatial information. We find that all layers contain representations of the grid state up to a linear transformation. Crucially, these representations are mostly dependent on how information is encoded in text, although we observe some degree of generalization, as SIRs from the same class produce more similar spatial encodings. We also find specific units that can represent the position of the agent in the grid by detecting its (x, y) coordinates, as well as detecting whether the agent is near the grid’s boundary. These units are remarkably general, detecting these properties regardless of SIR type, which indicates that they are not encoding prompt-specific features but the abstract concept of spatial location in the grid. Notably, these units are mostly located in intermediate and deeper layers, matching results from previous research that show higher-level reasoning and conceptual encodings in LLMs can be found in these layers~\cite{Geva2020,Durrani2022,Meng2022,Gurnee2023}. Put together, these findings point to inner models of space that are mostly format-dependent, with some elements encoding abstract spatial features, such as positions in 2D space.

Moreover, we demonstrate the existence of a subset of spatial units—also concentrated in intermediate layers—that can predict whether the selected action will get the agent closer to the goal, showing an understanding of spatial orientation regardless of how information is represented. Remarkably, these units are particularly active even during spatial reasoning tasks outside the grid-world context, suggesting that they can represent spatial information in a very abstract way, regardless of task and prompting. However, it’s important to note that these units may not exclusively process spatial information (i.e. they might not be monosemantic~\cite{Templeton2024,Bricken2023}). Furthermore, it is interesting that they are not strictly necessary to perform well on a spatial orientation task, as shown by our ablation experiments. This suggests that they are the most abstract elements of a spatial information processing system that appears to be widely distributed throughout the network. This is supported by the fact that the number of neurons whose activity correlates significantly with action correctness can reach tens of thousands across all model layers for certain SIRs, as shown in Fig.~\ref{fig4}. This wider representation of space appears to be less general (as it is SIR-specific) but sufficient to correctly navigate this simple 2D space. It also appears to be comprised of polysemantic neurons, which are necessary to preserve basic output coherence. In the future, it would be interesting to investigate whether the abstract components of the spatial model are necessary when solving harder problems or trying to navigate different kinds of environments. Furthermore, this model of space could be quite different in larger LLMs. We were not able to probe larger models due to constraints in memory and compute, but future work should look for equivalent structures in these larger LLMs to study how these internal representations of space change or stay invariant with scale.

Our work aims to enhance the explainability of LLM-based agentic systems by elucidating how the representation of spatial information influences model behavior. These insights have practical implications for the design of agentic systems. For instance, when deploying LLMs in environments where spatial reasoning is critical, explicitly formatting spatial data in a Cartesian manner can lead to more robust and efficient decision-making. Moreover, understanding the internal structure of spatial representations may inform future strategies for mechanistic analysis and interventions. These might involve reweighing or isolating identifiable units to mitigate unexpected behaviors, as well as using information encoded in these specific groups of neurons to make agentic decision-making more transparent, thus leading to safer and more reliable AI systems.

\appendix
\section{GWSOT Details}

An agent and a goal are placed inside a grid of size 5×5. The agent is positioned semi-randomly, ensuring that it starts at least two steps away from the goal. At each step of a trial, the agent must choose one of four possible actions—up, down, left, or right. All actions remain available even when the agent is at the grid’s boundary; if an action would move the agent outside the grid, its position is not updated. Prior to making a decision, the current world state is provided to the LLM as a user message. The full prompt given to the model is comprised of a conversation where user messages are world states and assistant messages represent past decisions. The prompt received by the LLM always starts with a user message (the initial world state) and alternates between user and assistant roles until a final user message (the current world state). This allows in-context learning, as the LLM can use information about how previous actions have altered the environment to adjust its strategy. The model is then required to output a JSON object containing a single field, "action", whose value is a string corresponding to the chosen move. Invalid responses result in the agent’s position not being updated, although these were only infrequently observed in the 1B and 3B models.

We define an action as correct if it brings the agent closer to the goal and as incorrect otherwise. A trial terminates when the agent successfully reaches the goal or when the maximum number of steps is exceeded, which is defined as 2×N (where N is the grid size). In the 5×5 grid, the agent is allowed a maximum of 10 steps. This number is chosen so that the agent can always reach the goal via a non-optimal path, even if the agent and the goal start positioned at opposite corners of the grid.

The following system prompt accompanies all representations to help the models understand the task instructions: 

\begin{lstlisting}[basicstyle=\ttfamily\small, breaklines=true, frame=single]
You are a navigation assistant tasked with guiding an Agent to a Goal in a 5x5 grid world. The Agent will start in a random position, and your objective is to provide directions that bring the Agent to the Goal.

Each time you receive an updated state of the world, choose the optimal next move to bring the Agent closer to the Goal. You may only respond with a JSON object containing a single field named "action", which should contain one of the following strings in all capital letters: "UP", "DOWN", "LEFT", or "RIGHT".

Ensure that your response is strictly in this format, with no additional text or commentary, since it will be used automatically and with no human supervision by a Python script to get the next world state.

Remember:
Your goal is to bring the Agent closer to the Goal as efficiently as possible.
Only respond with the JSON object in the exact format specified, using one of the four allowed action strings.
\end{lstlisting}

Examples of SIRs are provided in the right panels of Fig.~\ref{fig1}.

The full code to build and execute the GWSOT, as well as for all statistical analysis, is open source and available on \href{https://github.com/mneuronico/Griw-World-Spatial-Orientation-Task}{GitHub}.

\section{Summary of Experiments}

In this appendix, we provide a detailed summary of all experiments conducted for the Grid-World Spatial Orientation Task (GWSOT). For each experiment, we specify the models used, the spatial information representations (SIRs) applied, and the number of trials run under each condition. A random policy agent was also evaluated as a baseline. When analyzing experimental results, a statistical threshold of 0.05 was used throughout this study.

\subsection{Standard GWSOT Across Model Sizes}
We evaluated six LLaMA-3 models (3.2-1B, 3.2-3B, 3.1-8B, 3.2-11B, 3.1-70B, and 3.2-90B) on the standard 5×5 grid. For each model, every single SIR was tested individually.

\begin{table}[htbp]
\caption{Summary of Experiment 1}\label{tab:trials}
\setlength{\tabcolsep}{4pt}
\renewcommand{\arraystretch}{1.1}
\centering
\begin{tabular}{|l|c|}
\hline
Representation & Number of Trials per Model\\ \hline
JSON & 100\\ \hline
Chess Notation & 100\\ \hline
Symbol Grid & 100\\ \hline
Word Grid & 100\\ \hline
Row Description & 100\\ \hline
Column Description & 100\\ \hline
Random Policy & 100 (overall)\\ \hline
\end{tabular}
\end{table}

\subsection{Probing}
For probing analyses, we focused on the LLaMA-3.1-8B model using all SIR conditions. The final dataset comprised individual world-state/action pairs (between 3 and 10 per trial) along with their corresponding hidden-state activations.

\begin{table}[htbp]
\caption{Summary of Experiment 2}\label{tab:trials2}
\setlength{\tabcolsep}{4pt}
\renewcommand{\arraystretch}{1.1}
\centering
\begin{tabular}{|l|c|}
\hline
Representation & Number of Trials per Model\\ \hline
JSON & 50\\ \hline
Chess Notation & 50\\ \hline
Symbol Grid & 50\\ \hline
Word Grid & 50\\ \hline
Row Description & 50\\ \hline
Column Description & 50\\ \hline
\end{tabular}
\end{table}

\subsection{Ablation Studies}
We conducted ablation experiments on the LLaMA-3.1-8B model using only the JSON SIR. We silenced either units correlated with action correctness across all SIRs or units correlated with agent position across all SIRs. We also evaluated the effect of silencing all units correlated with action correctness only in JSON SIR trials, but this disrupted basic output coherence, so we only conducted a small number of trials for this condition.

\begin{table}[htbp]
\caption{Summary of Ablation Experiments}\label{tab:ablation}
\setlength{\tabcolsep}{4pt}
\renewcommand{\arraystretch}{1.1}
\centering
\begin{tabular}{|l|c|}
\hline
Ablation Condition & Number of Trials per Model\\ \hline
No Ablation (Control) & 100\\ \hline
Ablation of Core Action-Correctness Units & 100\\ \hline
Ablation of Core Agent-Position Units & 50\\ \hline
\parbox[t]{0.6\linewidth}{Ablation of All Units Correlated with JSON SIR Action-Correctness} & 10\\ \hline
\end{tabular}
\end{table}

\section{Linear Probing and Identification of Feature-Predicting Units}

In this appendix, we describe in detail the methods used to probe internal representations of spatial information in the LLaMA-3.1-8B model. Our analyses include (1) training linear regression models to decode the full grid configuration, (2) analyzing individual neurons for significant correlations with specific spatial features, (3) determining common significant units across different SIR types, and (4) silencing groups of units in ablation experiments while evaluating the model on a GWSOT.

\subsection{Linear Regression for Grid Configuration}

For each SIR, we first extracted activations from every layer of the model. Activations were averaged over all input tokens and then used to predict the complete configuration of the grid. Specifically, the target was a 50‑dimensional binary vector—25 dimensions indicating the presence ($1$) or absence ($0$) of the agent and 25 for the goal. For each SIR and for each layer separately, a standard linear regression model was trained on $90\%$ of the data (training set) and evaluated on the remaining $10\%$ (test set). A total of $(\mathrm{layers}\times\mathrm{SIRs}) = 32\times 6 = 192$ models were trained. The performance of these linear models was quantified by the coefficient of determination ($R^2$), both for the training and the test set. Training set $R^2$ was always $1.0$ for every model we trained.

To assess the invariance of the internal representations across different SIRs, we performed cross prediction experiments. Linear models trained on activations from one SIR were evaluated on the test sets of each of the other SIRs separately (i.e., each model was evaluated on six test sets, always receiving activations from the layer they were trained on). $R^2$ scores were computed for each evaluation and averaged across layers when applicable.

For each layer, we created a null baseline by shuffling the target values (i.e., permuting $Y_{\text{train}}$) to break the true relationship between activations and grid configuration. The null models consistently produced highly negative $R^2$ values (i.e., extremely poor predictive performance), confirming that the true models captured meaningful spatial information.

\subsection{Analysis of Individual Units}

To identify which units are predictive of specific spatial features, we fitted a simple linear model for each model parameter using each parameter’s activations when producing the first decision token to predict the spatial feature we cared about. In these analyses, the length of the predictor vector $X$ and the target vector $Y$ was the total number of world-state/action pairs in the dataset of a specific SIR type. $X$ was built from parameter activations in each step and $Y$ contained the spatial feature of interest (e.g., a binary indicator of whether the agent is in a particular grid location, or near the grid’s border, etc.). P-values for each unit were adjusted by using a Bonferroni correction to account for multiple comparisons (usually by $n_{\text{layers}} \times n_{\text{params\_per\_layer}}$, except when doing comparisons for every grid cell, where the number of comparisons was $5 \times 5 \times n_{\text{layers}} \times n_{\text{params\_per\_layer}}$). A significance threshold of $0.05$ was used to determine whether a specific parameter was significantly correlated with the spatial feature of interest after correction.

\subsection{Identification of Common Significant Units}

To isolate units that reliably encode spatial features regardless of the representation, we determined common units across all SIRs. For each SIR, a binary mask was created indicating whether a parameter’s corrected p-value (after Bonferroni adjustment) was below the significance threshold. Parameters that were significant for all SIRs were identified by taking the logical AND across the masks for all SIRs. We compare the number of parameters included in this mask with a shuffle control, where each layer’s parameter indices were randomly permuted between SIRs.

\subsection{Ablation Experiments}

In our ablation experiments, we silenced specific groups of units—identified as significantly correlated with spatial features—by applying a binary mask during the model’s forward pass. This procedure effectively zeroed out the contributions of the selected parameters while leaving the remainder of the network intact. We then evaluated the model on the 5×5 GWSOT, always using the JSON SIR, to compare its performance with and without ablation and computed the Success Rate for each condition.

\section{Additional Results from Probing Experiments}

\begin{figure}[!t]
\centering
\includegraphics[width=\textwidth]{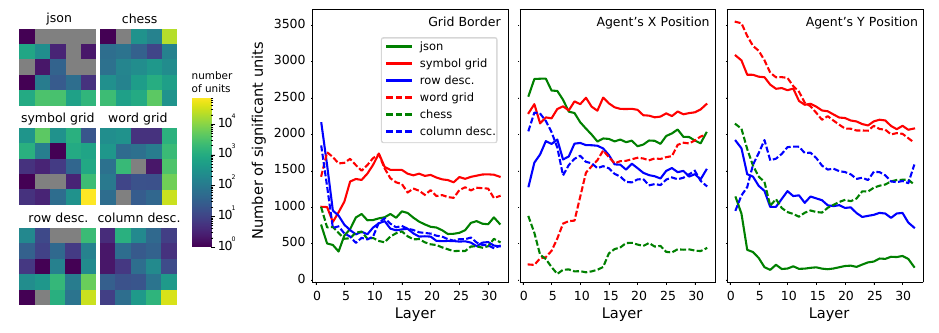}
\caption{Supplementary results from probing analysis. Left-most panel shows six heatmaps displaying the number of units that were significantly correlated with the position of the agent being a specific grid cell, for each SIR type. Gray cells denote places where no units were significantly correlated with that position. Note the color scale is logarithmic. The last three panels show the number of parameters per layer significantly correlated with a spatial feature of interest, for each SIR type. The features displayed are whether the agent is located on the grid border, the agent’s x coordinate and the agent’s y coordinate, respectively.} \label{fig5}
\end{figure}

In this appendix, we present supplementary analyses from our probing experiments on the LLaMA-3.1-8B model, focusing on the internal encoding of spatial information. The results summarized in Figure~\ref{fig5} provide further insight into how different SIRs influence the model’s activations.

First, when attempting to identify parameters correlated with specific grid locations, we observed that no individual unit was consistently correlated with the same grid cell across all SIR types. However, for single SIRs, we found that some grid positions were associated with thousands of significantly correlated units, while others had few or none (Fig.~\ref{fig5}, left-most panel, heatmaps). Although this pattern varied across different SIRs, it was relatively consistent when comparing SIRs within the same class (compare pairs of heatmaps in the same row). This might indicate differences in how the model processes the 2D environment, or it could simply reflect differences in the textual formulation of each SIR rather than a true abstraction of spatial features.

Furthermore, considering that each layer of the LLaMA model comprises 4096 parameters, it is notable that in certain layers and for specific SIRs more than half of these parameters show significant correlations with a spatial feature (last three panels of Fig.~\ref{fig5}). It is important to note that this large proportion of the model's parameters are likely correlated with the form of the input for each specific SIR (i.e. similar prompts should yield similar activations). If this is the case, the effect should be stronger in more superficial layers, whose representations are closer to the input. This is indeed what we see, in contrast to the distribution of parameters significantly correlated to the spatial feature of interest across all SIRs, which usually peaks in intermediate and deep layers. This suggests that these hundreds to thousands of significantly correlated parameters per layer for single SIRs are not truly encoding spatial information but just representing the input text.

\section{Alternative Spatial/Non-Spatial Reasoning Task}

\begin{table}[!t]
\caption{Example Prompt and Response Pairs for Alternative Task}\label{tab:prompts_responses}
\centering
\setlength{\tabcolsep}{4pt}
\renewcommand{\arraystretch}{1.1}
\begin{tabularx}{\linewidth}{|l|X|X|}
\hline
Condition & Prompt & Response \\ \hline
Spatial & You are in a library, you see a table to your left and a bookshelf to your right. You need to find the book which should be on the shelf next to the window. & To find the book, I would walk to the bookshelf next to the window, which is to my right. \\ \hline
Spatial & I'm in my office with my desk in front of me. To the right of the desk is a filing cabinet, and behind me is a bookshelf. I think I left my stapler near the filing cabinet; could you help me locate it? & Since the stapler is near the filing cabinet, and the filing cabinet is to the right of the desk, the stapler should be located to the right of the desk. \\ \hline
Non-Spatial & You are a hardworking person who has been told you often overwork yourself, and you wonder if that leads to burnout. & I often put in long hours, prioritizing work over self-care and relaxation. This constant push to be productive might be contributing to burnout. I've noticed I'm more irritable and exhausted lately, and my motivation has been wavering. Maybe it's time to reevaluate my work-life balance. \\ \hline
Non-Spatial & I live in the jungle, have a long tail, and often swing from trees. I enjoy eating bananas. What animal could I be? & You are likely a monkey. \\ \hline
\end{tabularx}
\label{tab4}
\end{table}

To investigate if units identified as processing spatial information in the GWSOT are also activated in an unrelated spatial context, we designed an additional task to compare spatial and non-spatial reasoning. We created a dataset of reasoning prompts that fall into two categories:

\begin{samepage} % Prevents page breaks within the environment
\begin{itemize}[label={\scalebox{1}{$\bullet$}}, leftmargin=*] % Change the bullet to a dot
    \item Spatial Reasoning Examples: These prompts describe everyday scenarios with explicit spatial cues (e.g., directions, relative positions) and require the model to reason about object locations.
    \item Non-Spatial Reasoning Examples: These prompts focus on abstract or personal reflections, descriptions, or general reasoning that do not emphasize spatial layout.
\end{itemize}
\end{samepage}

We used the following system prompt across all examples:

\begin{lstlisting}[basicstyle=\ttfamily\small, breaklines=true, frame=single]
You have been tasked with answering reasoning questions. Please answer precisely and briefly.
\end{lstlisting}

For each prompt, the model generated a response and we extracted corresponding hidden state activations for further analysis. Table~\ref{tab4} shows some examples of prompt/response pairs for the spatial and non-spatial conditions.

\begin{credits}
\subsubsection{\ackname} This study was funded by a PhD Scholarship from the National Scientific and Technical Research Council (CONICET).

\subsubsection{\discintname}
The authors have no competing interests to declare that are relevant to the content of this article.
\end{credits}

\end{document}